\documentclass[11pt,conference]{IEEEtran}
\IEEEoverridecommandlockouts
\usepackage{cite}
\usepackage{amsmath,amssymb,amsfonts}
\usepackage{algorithmic}
\usepackage{graphicx}
\usepackage{textcomp}
\usepackage{xcolor}
\def\BibTeX{{\rm B\kern-.05em{\sc i\kern-.025em b}\kern-.08em
    T\kern-.1667em\lower.7ex\hbox{E}\kern-.125emX}}

\usepackage{blindtext}
\usepackage{url}            %
\usepackage{booktabs}       %
\usepackage{amsfonts}       %
\usepackage{nicefrac}       %
\usepackage{microtype}      %
\usepackage{xcolor}         %

\usepackage{graphicx}
\usepackage{subfigure}
\usepackage{booktabs} %
\usepackage{dblfloatfix}
\usepackage{wrapfig}

\usepackage[pagebackref=true,breaklinks=true,letterpaper=true,colorlinks,bookmarks=false]{hyperref}

\usepackage{amsmath}
\usepackage{amsfonts}
\usepackage{booktabs}
\usepackage{bbm}
\usepackage{multirow}
\usepackage{multicol}
\usepackage{float}
\usepackage{placeins}
\usepackage[normalem]{ulem}
\usepackage{dcolumn}
\usepackage{siunitx}
\usepackage{makecell}

\def\f{f_{\theta}}
\def\e{e_{\phi}}
\def\L{\mathcal{L}}
\def\R{\mathbb{R}}

\begin{document}

\title{\vspace{10mm}Learning to Explain: A Model-Agnostic Framework for Explaining Black Box Models}

\author{Oren Barkan\thanks{*\hspace{0.2mm}Denotes equal contribution.}$^{\ast 1}$
\hspace{6mm} 
Yuval Asher$^{\ast2}$
\hspace{6mm}
Amit Eshel$^{\ast2}$
\hspace{6mm}
Yehonatan Elisha$^{1}$
\hspace{6mm}
\vspace{2mm}
Noam Koenigstein$^{2}$ 
\\
\vspace{2mm}
\normalsize{$^{1}$The Open University}
\hspace{7mm} \normalsize{$^{2}$Tel Aviv University}
}

\maketitle

\begin{abstract}
We present Learning to Explain (LTX), a model-agnostic framework designed for providing post-hoc explanations for vision models. The LTX framework introduces an ``explainer'' model that generates explanation maps, highlighting the crucial regions that justify the predictions made by the model being explained. To train the explainer, we employ a two-stage process consisting of initial pretraining followed by per-instance finetuning. During both stages of training, we utilize a unique configuration where we compare the explained model's prediction for a masked input with its original prediction for the unmasked input. This approach enables the use of a novel counterfactual objective, which aims to anticipate the model's output using masked versions of the input image. Importantly, the LTX framework is not restricted to a specific model architecture and can provide explanations for both Transformer-based and convolutional models. Through our evaluations, we demonstrate that LTX significantly outperforms the current state-of-the-art in explainability across various metrics. Our code is available at: \url{https://github.com/LTX-Code/LTX}
\end{abstract}


\section{Introduction and Related Work}
\label{sec:intro}

Over the past decade, the emergence of deep learning models has catalyzed significant advancements across a wide array of application domains, spanning natural language processing~\cite{Mikolov:Word13, barkan2017bayesian, vaswani2017attention, devlin2018bert, barkan2020scalable, liu2019roberta, barkan2020bayesian, ginzburg2021self, barkan2021representation}, recommender systems~\cite{he2017neural, wang2019neural, he2020lightgcn, malkiel2020recobert, barkan2019cb2cf, barkan2020attentive, barkan2020cold, barkan2021cold2, barkan2020neural, barkan2016item2vec, barkan2021anchor}, and audio processing~\cite{defossez2018sing, engel2019gansynth, barkan2019deep, kumar2019melgan, barkan2019inversynth, barkan2023inversynth}.
Notably, within the domain of computer vision, deep Convolutional Neural Networks (CNNs)~\cite{simonyan2014very, He2016DeepRL, Huang2017DenselyCC, Liu2022ACF}, in conjunction with more recent Vision Transformer (ViTs) models~\cite{dosovitskiy2020image}, has been marked by their exceptional performance across a diverse spectrum of vision-related tasks~\cite{imagenet, he2017mask, lu2019vilbert, carion2020end_detr, dai2016r,badrinarayanan2017segnet, Barkan_2023_discovery, he2022masked_mae}.

As a result, explainability methods have evolved as indispensable tools for unraveling the predictions of deep learning models across a spectrum of application domains~\cite{simonyan2013deep, chefer2021transformer, barkan2021grad, malkiel2022interpreting, gaiger2023not, barkan2020explainable, barkan2023modeling}. In the domain of computer vision, these explanation techniques are dedicated to crafting human interpretative aids in the form of heatmaps, shedding light on the crucial input regions that significantly impact the prediction of the vision model~\cite{zeiler2014visualizing, selvaraju2017grad,chefer2021transformer,barkan2021gam}.

Many earlier works have been focused on explaining CNNs. The two most dominant approaches are based on computing gradients w.r.t. the input or internal activation maps~\cite{sundararajan2017axiomatic,selvaraju2017grad,chattopadhay2018grad,barkan2021gam,jiang2021layercam,barkan2023dix,barkan2023six,Barkan_2023_ICCV}, and attribution methods~\cite{montavon2017explaining,shrikumar2017learning,nam2020relative}. Most relevant to this work, however, is a line of work in which a learning model is employed in order to produce explanations for the explained model. For example, in~\cite{chen2018learning}, a theoretical framework for instance-wise feature selection was presented for per-instance explanations. Another work by~\cite{fong2017interpretable} proposed extracting explanations by training a model to identify the image parts most influential in a classifier's decision. A key advantage of the learning approach is the ability to design optimization objectives according to varying requirements. Additionally, the option to select distinct monitoring metrics during training enhances the outcome's quality even further.

There are relatively fewer works focusing on explaining Vision Transformers (ViTs)~\cite{dosovitskiy2020image}. Early attempts primarily involved leveraging attention scores. However, the aggregation of attention scores from different layers poses an ongoing research challenge. For instance, simple averaging has been observed to induce blurring, and researchers have explored various heuristics to address this issue~\cite{abnar2020quantifying}.
Transformer Attribution (T-Attr)~\cite{chefer2021transformer} assigns local relevance based on a Deep Taylor Decomposition (DTD) and propagates it forward through the layers. Their approach accounts for skip-connections which have challenged previous methods. The authors later generalized this method and presented Generic Attention Explainability (GAE) for explaining Bi-Modal transformers~\cite{chefer2021generic}.

In this work, we present Learning To Explain (LTX) - a model-agnostic framework for explaining vision models.
The LTX framework involves the development of an \emph{explainer} model which is trained to produce \emph{explanation maps} highlighting areas in the input that best explain the predictions of the \emph{explained} model. As such, the explainer model is a trained self-supervised prediction model that learns to identify areas in the input image that were most influential in the explained model's prediction process.

The LTX framework is agnostic to the explained model's architecture and can be applied to any vision model. Unlike non-agnostic explanation methods, LTX does not require access to the internal parameters of the explained model, and its sole prerequisite is the ability to derive gradients w.r.t. the explained model's input. 
The training of the explainer model takes place in two steps. The first is a pretraining phase, where the explainer model learns to create explanation maps that highlight the most influential areas in the input image for the predictions of the explained model. This is followed by a per-instance finetuning phase, where the explainer model is further refined to improve its explanation of a particular sample. In addition, LTX's finetuning enables class-specific maps that explain the model's prediction w.r.t. any class.

LTX's optimization is built around a unique self-supervised counterfactual prediction task akin to counterfactual perturbation tests. 
As we show later, this necessitates the inclusion of the explained model within the explainer's optimization objective which enables the propagation of gradients from the explained model to the explainer. In our evaluations, we applied the LTX framework to ViTs and CNNs, achieving state-of-the-art results on both model architectures.

\section{The Learning to Explain Framework}
\label{sec:method}

LTX employs an \emph{explainer-explained} abstraction in which an explainer model $e$ is trained to explain the predictions made by the explained model $f$.
Let $\f:[0,1]^n\rightarrow\ [0,1]^k$ be the \emph{explained} model. W.l.o.g, we assume $\f$ is a classifier emitting a discrete probability distribution over $k$ classes. Hence $\f$, parameterized by $\theta$, receives an instance $x\in \R^n$ as input and generates a prediction $y\in \R^k$. 

An \emph{explanation map} is a vector $m\in [0,1]^n$ scoring the attribution of each element in $x$ to the prediction $\f(x)$. Namely, $m_i$, the $i$-th element in $m$, quantifies the responsibility of element $x_i$ on the prediction $\f(x)$.
Given an explanation map $m$, we create a \emph{masked} version of $x$ denoted by $x^m:=\psi(x,m)$. In this masked version, the elements that best explain the prediction are preserved, while the less relevant elements are obscured. To this end, a higher explanation score $m_i$ implies a lower masking effect on $x_i$, hence retaining more of the original information of $x_i$. In contrast, a high masking of $x_i$ replaces the original value of $x_i$ with an uninformative value.

The function $\psi$, which serves as a masking mechanism, can be realized differently based on the specific domain of application and the nature of $\f$. For instance, in vision applications, $\psi$ can blur regions in $x$ that are associated with low explanation scores in $m$. A straightforward but debatable method might involve merging $x_i$ with a fixed value, e.g., blackening. However, blackening isn't always an effective masking technique, as black pixels aren't altered in this approach. Another option is to blend $x_i$ with a random value from its empirical distribution, making $\psi$ stochastic. If $\f$ supports incomplete inputs (e.g., $\f$ encounters masked examples during the training phase), the same mechanism can be utilized for the masking operation. For example, in Transformer-based models, special tokens such as [MASK] or [PAD] can be employed to mask tokens in the input. Another option involves simulating missing information via a dropout-like mechanism. Finally,  it's possible to train a specific variable to signify masked values. Depending on the type of input space, this variable could take the form of a tensor, vector, or scalar. For instance, in the realm of NLP, this might take the shape of a unique token (vector) that is combined with the original token to achieve masking.

In our study, we chose the straightforward masking method based on a constant value, training a scalar parameter $z\in \R$  that is combined to mask elements in the input, similar to the concept of blackening. Despite its potential limitations explained above, this approach has shown to be effective in practice. Hence, $\psi$ employs a linear blending as follows: 
\begin{equation}
\label{eq:psi}
x^m:=\psi(x,m)=x \circ m + z(1 - m). 
\end{equation}
An \emph{explainer} is a parametric function $\e: \R^n \times \R^n \rightarrow [0,1]^n$, parameterized by $\phi$, that receives an instance $x$, a prediction (or label) $y$, and produces an explanation map $m=\e(x, y)$. The explainer parameters $\phi$ are learned w.r.t. the inputs and the corresponding predictions produced by $\f$ through the LTX optimization process. Importantly, while the explained model $\f$ actively participates in optimizing the explainer $\e$, its weights are already optimized and kept frozen throughout the entire optimization process. The sole requirement for $\f$ is the calculation of its gradients w.r.t. its input. As we explain next, LTX facilitates a counterfactual loss that enables implicit information flow regarding $y$ via $\f$ back to the explainer $\e$. Hence, the explicit provision of $y$ as input to $\e$ is optional. In other words, as long as $\e$ is exposed to information regarding $y$ through $\f$, it is sufficient to supply $x$ as a single input to $\e$, and having $m=\e(x)$.

\begin{figure*}
    \begin{center}
  \includegraphics[width=\textwidth]{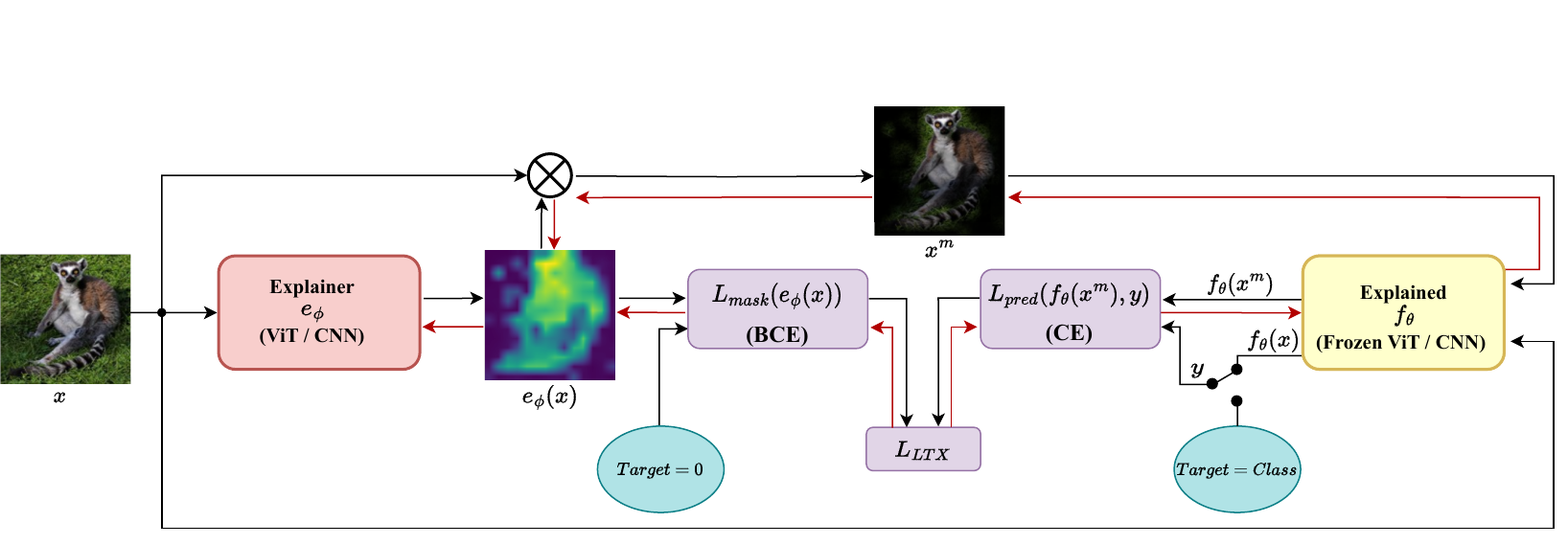}\vspace{-8mm}
  \caption{The LTX framework: The \emph{explainer} model $\e$ is trained to explain the predictions of the \emph{explained} model $\f$ based on prediction loss on the masked input $\L_{pred}$ and the regularization term $\L_{mask}$ (in this work, $\L_{inv}$ and $\L_{smooth}$ from Eq.~\ref{eq:lte-loss} do not participate in the optimization). While the explained model $\f$ takes an active part in the optimization of $\e$, its weights are kept frozen throughout the LTX optimization process. The red lines illustrate gradient propagation.} \vspace{-1mm}
\label{fig:illustration_of_LTE_implementation_for_VIT}
  \end{center}
\end{figure*}

\subsection{The LTX Objective}
\label{sec:LTE}
A good explainer is expected to produce an explanation map that highlights elements in $x$ crucial for explaining the prediction $\f(x)$. Therefore, the explainer $\e$ is optimized to mask as much as possible from the original input, while retaining the essential elements in $x$, ensuring that the prediction on the masked instance $\f(x^m)$ closely aligns with the prediction on the original instance $\f(x)$. To this end, we introduce the LTX loss:
\begin{equation}
\label{eq:lte-loss}
\begin{split}
 \L_{LTX}(x,y)&=\L_{pred}(\f(x^m), y) 
 + \lambda_{mask}\L_{mask}(m) \\&+ \lambda_{inv}\L_{inv}(\f(x^{1-m}), y) \\&+\lambda_{smooth}\L_{smooth}(m),
\end{split}
\end{equation}
where $x^m$ is defined in Eq.~\ref{eq:psi}, and $m=\e(x)$ is the explanation map produced by the explainer. The optimization of $\L_{LTX}$ is performed on the explainer parameters $\phi$ and the masking parameter $z$, while $\f$ remains frozen throughout the optimization process. Next, we elucidate the role of each loss term in Eq.~\ref{eq:lte-loss}.

The first loss term $\L_{pred}$ measures the discrepancy between the prediction for the masked input $x^m$ and the label $y$. A natural choice is setting $\L_{pred}$ to the cross-entropy (CE). Notably, $y$ can take several values: (1) $\f(x)$, the original distribution over the $k$ classes based on the unmasked input $x$. (2) One-hot vector representing the class with the highest probability in $\f(x)$. (3) One-hot vector based on the user-specified target class. The second term, $\L_{mask}$, is a regularization term that encourages sparse and narrow explanation maps. The second term, $\L_{mask}$ is a regularization term encouraging sparse and narrow explanation maps. A possible choice for $\L_{mask}$ is the $L1$ loss or the binary cross-entropy (BCE) loss with the zero vector. The third and fourth terms are optional. $\L_{inv}$ is a counterfactual loss encouraging the prediction for the inversely masked input to diverge from $y$, using the inverse of the explanation map, i.e., $1-m$. Possible choices for $\L_{inv}$ are the negative entropy of $\f(x)$ (effectively maximizing the entropy of the prediction), the actual probability score for the class $y$, or $-log(1-\f(x^{1-m})_{y})$. $\L_{smooth}$ encourages smooth explanation maps. Possible choices are the L1 or L2 total variation. Lastly, it is worth noting that instead of soft masking (Eq.~\ref{eq:psi}), one can apply hard masking. This can be implemented either by straightforward thresholding or by introducing a Bernoulli parameter for each entry in the mask and sampling the mask during optimization, utilizing the Gumbel-Softmax trick.

In this work, we employed LTX using the first two loss terms from Eq.~\ref{eq:lte-loss}, i.e., $\L_{pred}$ and $\L_{mask}$, leaving the investigation of the last two terms for future research). Specifically, we set $\L_{pred}$ to the the CE, and $\L_{mask}$ to the BCE with the zero label (our experimentation shows that both BCE and L1 perform similarly). Finally, the resulted implementation of LTX in this work is illustrated in  Fig.~\ref{fig:illustration_of_LTE_implementation_for_VIT}.

\subsection{LTX Pretraining and Finetuning}
\label{sec:lte-pretraining-finetuning}
LTX is based on a pretraining phase followed by a per-instance finetuning phase. 
In the pretraining phase, the explainer $\e$ is optimized on a dataset $X=\{x_i\}_{i=1}^N$ that contains representative instances from the classes supported by the explained model $\f$. Optimization proceeds with stochastic gradient descent on $\phi$ and $z$ w.r.t. $\frac{1}{N}\sum_{i=1}^N \L_{LTX}(x_i,y_i)$.

Empirically, we found that initializing the explainer with a pretrained backbone, similar to that of the explained model, is effective when pretraining the explainer on a relatively small dataset. For example, for an explained ViT model that was pretrained on ImageNet, we were able to obtain a good pretrained explainer $e_{\theta_{X}}$ by first initializing it with a clone of the same trained ViT backbone as the explained model, and then pretraining it using only a single instance per class i.e., 1000 instances. Yet, it is important to clarify that LTX supports explanations for any black box model, and the explainer's architecture does not necessarily need to match that of the explained model.

In the finetuning phase, LTX further refines the explainer's parameters per a specific instance, enhancing its ability to provide a tailored explanation. When presented with an instance $x$, we conduct finetuning on the explainer using $\L_{LTX}$ (Eq.~\ref{eq:lte-loss}) to obtain the instance-specific explainer $e_{\theta_{x}}$. Subsequently, the final explanation map is computed as $m=e_{\theta_{x}}(x)$. A significant advantage of LTX lies in the ability to assess the explainer's performance on a designated validation set w.r.t. a specific metric of interest. One can then choose the explainer checkpoint that performs optimally, both during the pretraining and finetuning phases. Finally, it's important to clarify that the finetuning process is designed to result in improvement; if not, we revert to using the explanation map produced by the pretrained explainer.

\subsection{LTX Implementation for ViTs and CNNs}
\label{sec:ltx-impl}
So far, we have explained the LTX framework in a general form. In what follows, we provide a concrete description of the specific explainer architectures we employed for explaining ViT and CNN models. 

\subsubsection{Explaining ViTs}
\label{sec:lte-vit}
ViT analyzes images at a patch level. Given an input image $x$, ViT divides it into $n$ non-overlapping patches that are transformed into a sequence of $q$-dimensional token representations. A special \textsc{[CLS]} representation is prepended to the sequence and positional embeddings are added to each token.
These tokens are propagated via a stack of encoders that outputs a new sequence of representations $H(x):=(h_j(x))_{j=1}^n$, and a new \textsc{[CLS]} representation $h_{CLS}(x)$. Finally, a MLP classifier is applied on top of $h_{CLS}(x)$ to produce the final prediction. We refer the reader to~\cite{dosovitskiy2020image} for a detailed description of the ViT architecture.

The explainer model $\e$ for ViTs starts from a clone of the trained explained model $\f$. 
However, unlike $\f$ which uses $h_{CLS}(x)$ for its predictions, $\e$ applies a shared MLP classifier $v$ on top of each token representation from $H$ in order to predict the explanation score for each corresponding patch.
Namely, $\e(x)=v(H(x))=[v(h_1(x)),...,v(h_d(x))]^T$ is a vector in $[0,1]^n$ that forms a patch-level map where $v(h_i(x))$ is the explanation score for the $i$-th patch. 
In this work, $v$ is implemented as a $768\rightarrow768\rightarrow1$ MLP with a single hyperbolic tangent activated hidden layer, and a sigmoid activation on the output. Finally, we interpolate the 196 explanation scores assigned to the 196 patches (each corresponding to a 16x16 region) to the pixel-level of the original image dimensions (224x224) in order to get the final explanation map.

\subsubsection{Explaining CNNs}
\label{sec:lte-cnn}
Similar to ViTs, we set the initial explainer $\e$ as a replica of the backbone of the explained model $\f$. Specifically, we extract only the convolutional blocks from $\f$, discarding all subsequent fully connected layers, including the classification head. As a result, the last layer outputs a 3D tensor with spatial dimensions $r \times r$ and $c$ channels. Subsequently, a $1\times1$ convolution is applied, followed by a sigmoid activation, yielding a 2D tensor of size $r \times r$. This tensor is then resized using bilinear interpolation to generate the final explanation map that matches the spatial dimension of the original input. Note that a more sophisticated approach could involve learning a U-Net based decoder rather than employing aggressive interpolation. However, this investigation is left for future research.

\section{Experimental Setup and Results}
\label{sec:experiments}

Our evaluations cover explanations for ViT and CNN models. For ViTs we consider ViT-Base (\textbf{ViT-B})~\cite{dosovitskiy2020image} from HuggingFace\footnote{\url{https://huggingface.com/}}. 
For CNNs we consider ResNet101 (\textbf{RN})~\cite{he2016deep}. The experiments were conducted on the ImageNet ILSVRC 2012 (\textbf{ImageNet})~\cite{imagenet} validation consisting of 50K images from 1000 classes.  %
\subsection{Evaluation Tasks and Metrics}
We followed the protocol from~\cite{chefer2021transformer}, which is the current state-of-the-art in explaining ViTs, and report the Negative Perturbation AUC (\textbf{NEG}) and the Positive Perturbation AUC (\textbf{POS}). \textbf{NEG} is a counterfactual test that entails a gradual blackout of the pixels in the original image in increasing order according to the explanation map while searching to see when the model's top predicted class changes. By masking pixels in increasing order,  we expect to remove the least relevant pixels first, and the model's top predicted class is expected to remain unchanged for as long as possible. Results are measured in terms of the Area Under the Curve (AUC), and higher values are considered better. Accordingly, the \textbf{POS} test entails masking the pixels in decreasing order with the expectation that the model's top predicted class will change quickly, hence in \textbf{POS}, lower values are better. 

In addition, we followed~\cite{petsiuk2018rise} and report the Insertion AUC (\textbf{INS}) and Deletion AUC (\textbf{DEL}) perturbation tests. 
\textbf{INS} and \textbf{DEL} entail a gradual blackout in increasing or decreasing order, similar to \textbf{NEG} and \textbf{POS}, respectively. 
However instead of tracking the point at which the top predicted class changes, in \textbf{INS} and \textbf{DEL} the AUC is computed w.r.t. the predicted probability of the top class. By masking pixels according to increasing (decreasing) order of importance, we expect that the predicted probability of the top class will decrease slowly (quickly). Hence, for \textbf{INS} higher values are better and for \textbf{DEL} lower values are better.

\subsection{LTX Setup}
The pretraining phase utilizes a random sample of 2000 images from the ImageNet training set. This sample is evenly divided into training and validation sets, denoted as $X$ and $X_{val}$, respectively, with each set containing a single example per class.
As explained in Sec.~\ref{sec:lte-pretraining-finetuning}, one advantage of LTX is its ability to monitor per specific metric either during the pretraining phase (on the validation set), and / or the finetuning phase (on a single example), and select the explainer model that performs the best on this metric. In this work, we only monitored the POS and NEG metrics. Hence, further improvements may be possible for the other metrics.  %

Recall that the LTX loss can be optimized w.r.t. the predicted class, the ground-truth class, or the prediction vector. In this work, we report results for the predicted (\textbf{P}) and the ground-truth target (\textbf{T}) classes. %

We used the Adam optimizer~\cite{kingma2014adam} with learning rate 2e-3, batch size 32, and set $\lambda_{mask}=30$ in Eq.~\ref{eq:lte-loss} (recall that in this work, $\L_{inv}$ and $\L_{smooth}$ do not participate in the optimization).
The instance-specific finetuning phase was controlled by a hyperparameter that sets the maximum number of update steps when not reaching convergence through the monitored metric. Hence, we set the maximum number of update steps to 25.

\subsection{Evaluated Methods}
We consider different state-of-the-art methods for explaining ViTs and CNNs.
Baselines for explaining ViTs include: 
\begin{itemize}
    \item \textbf{rollout} - the rollout method~\cite{abnar2020quantifying} that utilizes the attention maps from all layers in the transforme
    \item \textbf{raw-att} - the `raw attention' baseline that was used in~\cite{chefer2021transformer}. The raw-att method utilizes the attention map from the last attention layer in the transformer to extract the relevance scores.
    \item \textbf{GC} - an adaptation of the Grad-CAM method for transformers as proposed by~\cite{chefer2021transformer}.
    \item \textbf{pLRP} - the partial LRP method~\cite{voita2019analyzing}, which is a modified version of LRP~\cite{binder2016layer} that utilizes attention information in the last encoder layer.
    \item \textbf{T-Attr} - the state-of-the-art Transformer Attribution method~\cite{chefer2021transformer} that assigns local relevance based on a Deep Taylor Decomposition (DTD) and propagates it forward through the layers.
    \item \textbf{GAE} - the Generic Attention Explainability method~\cite{chefer2021generic}, which is a state-of-the-art method for explaining Bi-Modal Transformers.
    
\end{itemize}

For CNNs, we include the following baselines:
\begin{itemize}
\item \textbf{GC} - the Grad-CAM method~\cite{selvaraju2017grad}. %
\item \textbf{GC++} - Grad-CAM++ from~\cite{chattopadhay2018grad}, which is an improvement of the original Grad-CAM method. %
\item \textbf{XGC} - XGrad-CAM~\cite{Fu2020AxiombasedGT} computes activation coefficients based on axioms related to the operation of the CNNs. %
\item \textbf{LIFT} - LIFT-CAM~\cite{Jung2021TowardsBE} employs the DeepLIFT technique to estimate activation maps based on SHAP values~. %
\item \textbf{AC} - Ablation-CAM~\cite{Desai2020AblationCAMVE} is a gradient-free method that scores the activations by masking them and measuring their relative importance.
\item \textbf{FG} - FullGrad~\cite{Srinivas2019FullGradientRF} considers gradient w.r.t. the bias terms, beyond the input.
\item \textbf{LC} - LayerCAM~\cite{jiang2021layercam} generates fine-grained object localization information from class activation maps.
\end{itemize}

\begin{table*}[t]
     \caption{Explaining VIT-B w.r.t. the predicted (P) and target (T) classes. For NEG and INS (POS and DEL) higher (lower) is better.}
  \vspace{-5mm}
   \begin{center}
       \scalebox{1.1}{
    \begin{tabular}{@{}lc@{}lc@{}lc@{}lc@{}lc@{}lc@{}lc@{}}%
    \toprule
      & & \multicolumn{1}{l}{rollout} &    \multicolumn{1}{l}{raw-att} & \multicolumn{1}{l}{GC} & \multicolumn{1}{l}{pLRP} &     \multicolumn{1}{l}{T-Attr} & \multicolumn{1}{l}{GAE} & \multicolumn{1}{l}{LTX} \\
    \midrule
    \multirow{2}{*}{NEG $\uparrow$} & \multicolumn{1}{l}{P} & \multicolumn{1}{l}{53.10}  & \multicolumn{1}{l}{45.55}  & \multicolumn{1}{l}{41.52}  & \multicolumn{1}{l}{50.49}  & \multicolumn{1}{l}{54.16} & \multicolumn{1}{l}{\uline{54.61}} & 
     \multicolumn{1}{l}{\textbf{64.83}} \\
         & \multicolumn{1}{l}{T} & \multicolumn{1}{l}{53.10} & \multicolumn{1}{l}{45.55} & \multicolumn{1}{l}{42.02}  & \multicolumn{1}{l}{50.49}  & \multicolumn{1}{l}{55.04} & \multicolumn{1}{l}{\uline{55.67}} & \multicolumn{1}{l}{\textbf{65.49}} \\
    \multirow{2}{*}{POS $\downarrow$} & \multicolumn{1}{l}{P} & \multicolumn{1}{l}{20.05}  & \multicolumn{1}{l}{23.99} & \multicolumn{1}{l}{34.06} & \multicolumn{1}{l}{19.64} & \multicolumn{1}{l}{\uline{17.03}} & 
    \multicolumn{1}{l}{17.32} & \multicolumn{1}{l}{\textbf{11.19}} \\
         & \multicolumn{1}{l}{T} & \multicolumn{1}{l}{20.05} & \multicolumn{1}{l}{23.99} & \multicolumn{1}{l}{33.56} & \multicolumn{1}{l}{19.64} &
         \multicolumn{1}{l}{\uline{16.04}} &
         \multicolumn{1}{l}{{16.72}} & \multicolumn{1}{l}{\textbf{10.33}} \\

    \multirow{2}{*}{INS $\uparrow$} & \multicolumn{1}{l}{P} & \multicolumn{1}{l}{47.27}  & \multicolumn{1}{l}{40.11} & \multicolumn{1}{l}{34.94} & \multicolumn{1}{l}{44.61} & \multicolumn{1}{l}{48.58} & 
    \multicolumn{1}{l}{\uline{48.96}} & \multicolumn{1}{l}{\textbf{54.40}} \\
         & \multicolumn{1}{l}{T} & \multicolumn{1}{l}{47.27} & \multicolumn{1}{l}{40.11} & \multicolumn{1}{l}{35.24} & \multicolumn{1}{l}{44.61} &
         \multicolumn{1}{l}{49.19} &
         \multicolumn{1}{l}{\uline{49.65}} & \multicolumn{1}{l}{\textbf{54.30}} \\

    \multirow{2}{*}{DEL $\downarrow$} & \multicolumn{1}{l}{P} & \multicolumn{1}{l}{16.79}  & \multicolumn{1}{l}{20.07} & \multicolumn{1}{l}{28.59} & \multicolumn{1}{l}{16.53} & \multicolumn{1}{l}{\uline{14.20}} & 
    \multicolumn{1}{l}{14.37} & \multicolumn{1}{l}{\textbf{11.07}} \\
         & \multicolumn{1}{l}{T} & \multicolumn{1}{l}{16.79} & \multicolumn{1}{l}{20.07} & \multicolumn{1}{l}{28.28} & \multicolumn{1}{l}{16.53} &
         \multicolumn{1}{l}{\uline{13.77}} &
         \multicolumn{1}{l}{{13.99}} & \multicolumn{1}{l}{\textbf{10.72}} \\
    \bottomrule
    
  \end{tabular}}
  \end{center}\vspace{-1mm}
  \label{tab:vit_base_perturbation_test_table}
\end{table*}
\begin{table*}[t]
 \begin{center}
 \caption{Explaining RN w.r.t. the predicted (P) and target (T) classes. For NEG and INS (POS and DEL) higher (lower) is better.}
 \label{tab:table_cnns}
    \vspace{-3mm}
   \scalebox{1.1}{
    \begin{tabular}{@{}lc@{}lc@{}lc@{}lc@{}@lc@{}lc@{}lc@{}}%
    \toprule
      & & & \multicolumn{1}{l}{GC} & \multicolumn{1}{l}{GC++} & \multicolumn{1}{l}{XGC} & \multicolumn{1}{l}{LIFT} & \multicolumn{1}{l}{AC} & \multicolumn{1}{l}{FG} & \multicolumn{1}{l}{LC} & \multicolumn{1}{l}{LTX} \\
    \midrule
    
    & \multirow{2}{*}{NEG $\uparrow$} & \multicolumn{1}{l}{P} & \multicolumn{1}{l}{\uline{55.78}} & \multicolumn{1}{l}{55.55} & \multicolumn{1}{l}{55.41} & \multicolumn{1}{l}{{55.60}} & \multicolumn{1}{l}{53.27} & \multicolumn{1}{l}{54.73} & \multicolumn{1}{l}{{55.05}} & \multicolumn{1}{l}{\textbf{61.13}} \\
    
         & & \multicolumn{1}{l}{T} & \multicolumn{1}{l}{\uline{55.60}} & \multicolumn{1}{l}{55.88} & \multicolumn{1}{l}{55.21} & \multicolumn{1}{l}{55.62} & \multicolumn{1}{l}{52.26} & \multicolumn{1}{l}{54.34} & \multicolumn{1}{l}{54.99} & \multicolumn{1}{l}{\textbf{62.80}} \\
    
    & \multirow{2}{*}{POS  $\downarrow$} & \multicolumn{1}{l}{P} & \multicolumn{1}{l}{13.50} & \multicolumn{1}{l}{13.92} & \multicolumn{1}{l}{{13.54}} & \multicolumn{1}{l}{14.84} & \multicolumn{1}{l}{13.02}  & \multicolumn{1}{l}{\uline{12.31}} & \multicolumn{1}{l}{13.84} & \multicolumn{1}{l}{\textbf{8.65}} \\
         
         & & \multicolumn{1}{l}{T} & \multicolumn{1}{l}{\uline{13.32}} & \multicolumn{1}{l}{13.55} & \multicolumn{1}{l}{13.66}  & \multicolumn{1}{l}{13.46} & \multicolumn{1}{l}{14.81} & \multicolumn{1}{l}{13.67} & \multicolumn{1}{l}{13.91} & \multicolumn{1}{l}{\textbf{8.79}} \\
    
    & \multirow{2}{*}{INS $\uparrow$} & \multicolumn{1}{l}{P} & \multicolumn{1}{l}{\textbf{51.50}} & \multicolumn{1}{l}{51.09} & \multicolumn{1}{l}{51.04} & \multicolumn{1}{l}{\uline{51.18}} & \multicolumn{1}{l}{49.78} &  \multicolumn{1}{l}{50.33} & \multicolumn{1}{l}{50.72} & \multicolumn{1}{l}{50.52} \\
         
         & & \multicolumn{1}{l}{T} & \multicolumn{1}{l}{\textbf{52.47}} & \multicolumn{1}{l}{51.69} & \multicolumn{1}{l}{51.63} & \multicolumn{1}{l}{\uline{51.90}} & \multicolumn{1}{l}{49.79} & \multicolumn{1}{l}{50.75} & \multicolumn{1}{l}{51.25} & \multicolumn{1}{l}{51.37} \\
    
    & \multirow{2}{*}{DEL $\downarrow$} & \multicolumn{1}{l}{P} & \multicolumn{1}{l}{\uline{12.15}} & \multicolumn{1}{l}{12.25} & \multicolumn{1}{l}{12.25} & \multicolumn{1}{l}{12.16} & \multicolumn{1}{l}{14.60} & \multicolumn{1}{l}{12.36} & \multicolumn{1}{l}{12.63} & \multicolumn{1}{l}{\textbf{9.24}} \\
         
         & & \multicolumn{1}{l}{T} & \multicolumn{1}{l}{\uline{12.13}} & \multicolumn{1}{l}{12.24} & \multicolumn{1}{l}{12.22} & \multicolumn{1}{l}{12.81} & \multicolumn{1}{l}{13.60} & \multicolumn{1}{l}{13.12} & \multicolumn{1}{l}{12.36} & \multicolumn{1}{l}{\textbf{9.45}} \\

        \bottomrule
      \end{tabular}}\vspace{-1mm}
  \end{center}      
      
    \end{table*}

\subsection{Results}
\label{sec:results}

Table~\ref{tab:vit_base_perturbation_test_table} presents the explanation results for ViT-B according to the POS and NEG metrics, on the predicted and target (ground-truth) classes on the ImageNet dataset~\cite{imagenet}. Arguably, evaluations on the predicted class are more akin to the task of explanations than evaluations based on the target class. Regardless, we cover both for the sake of compatibility with state-of-the-art baseline e.g., \cite{chefer2021transformer,chefer2021generic}. 
Note that rollout, raw-attn, and pLRP are class-agnostic methods, hence producing the same maps for the target and the predicted classes.
Table~\ref{tab:vit_base_perturbation_test_table} indicates that LTX outperforms all other methods by a significant margin. %

Table~\ref{tab:table_cnns} presents the results for CNN models. LTX demonstrates superior performance on the NEG, POS, and DEL tests, while remaining competitive on INS. As discussed in Sec.~\ref{sec:ltx-impl}, there is potential for enhancing LTX, especially for CNNs, by exploring improved upsampling techniques. This consideration arises from the fact that, in the case of RN, the explainer generates low resolution $7\times 7$ explanation maps. In this scenario, upsampling through interpolation back to the original image dimensions ($224\times 224$) may be overly aggressive. This is in contrast to ViT-based explainer that produces higher resolution maps ($14\times 14$). Therefore, we believe that employing more sophisticated upsampling techniques, such as learning designated decoders, has the potential to yield better explanation maps with LTX. Lastly, as mentioned earlier, in this work, we monitored the validation set for NEG and POS. Specifically, for DEL and INS, we report results based on the best-performing checkpoint w.r.t. POS. Thus, individual monitoring of DEL and INS may lead to further gains.

\begin{figure*}[t]
  \centering
  \includegraphics[scale=1]{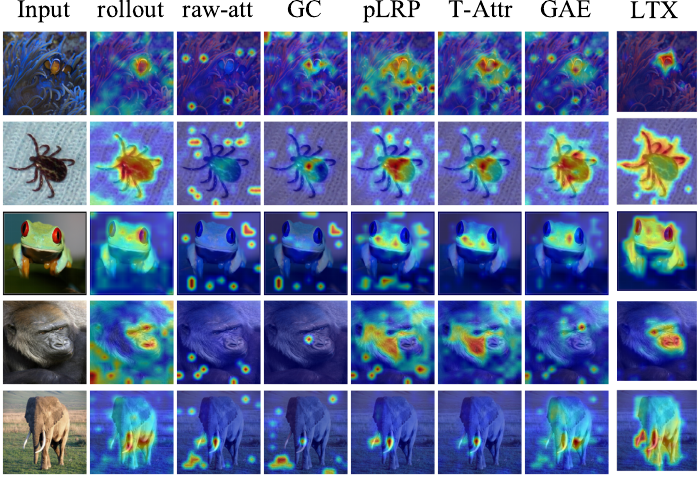}\vspace{-3mm}
  \caption{Comparison of ViT-B explanation maps for single-class images. %
  }
    \label{fig:single_class_vis}
\end{figure*}

\begin{figure*}[t]
  \centering
    \includegraphics[scale=1]
    {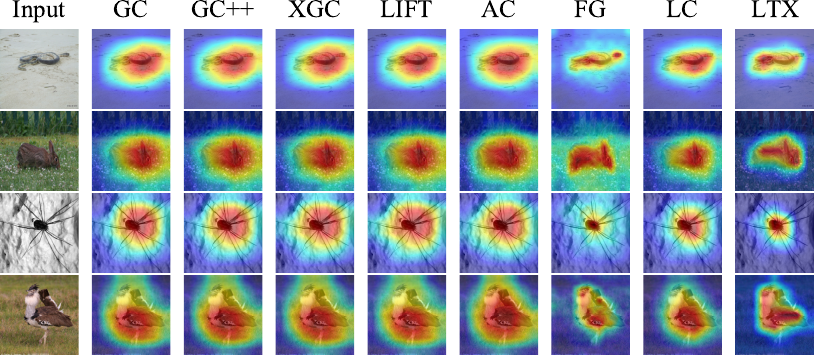}\vspace{-3mm}
    \caption{Comparison of RN explanation maps for single-class images. %
    }\vspace{-3mm}
    \label{fig:cnn_single_class_vis_small}
\end{figure*}

\begin{figure*}[t]
  \centering
    \includegraphics[scale=1]
    {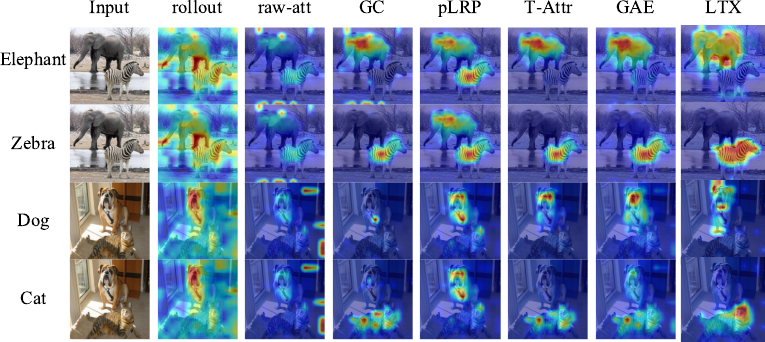}\vspace{-3mm}
    \caption{Comparison of explanation maps in multi-class images.}
    \label{fig:2_classes_vis}
    \end{figure*}

\begin{figure*}[t]
    \centering
        \includegraphics[scale=1]{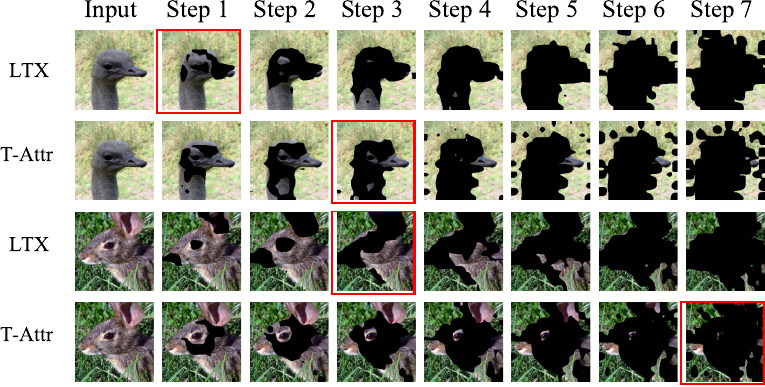}
    \caption{LTX vs. T-Attr~\cite{chefer2021transformer} POS perturbation steps: The red square indicates the step in which the prediction changes.}\vspace{-1mm}
    \label{fig:perturbation_tests_steps_comparison}
\end{figure*}

\begin{figure}[t]
  \centering
    \centering
        \includegraphics[width=\columnwidth]{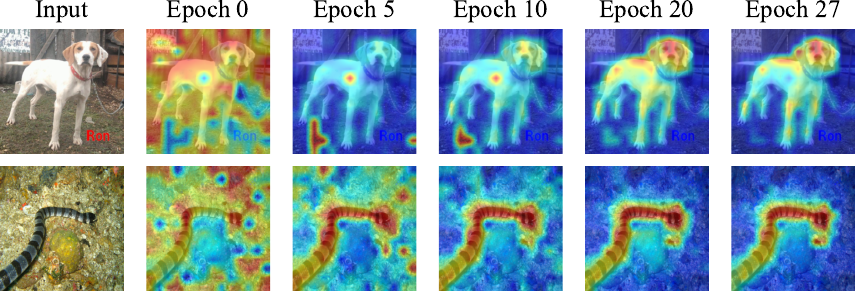}\vspace{-7mm}
    \caption{Convergence across the pretraining epochs of LTX.}
    \label{fig:training_procedure_examples}
\end{figure}

\begin{figure}[t]
    \centering
    \includegraphics[width=\columnwidth]{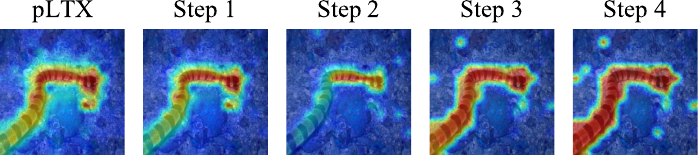}\vspace{-7mm}
    \caption{LTX finetuning example.}\vspace{-5mm}
    \label{fig:LTE_process_from_pre_LTE}
\end{figure}

Figures~\ref{fig:single_class_vis}-\ref{fig:cnn_single_class_vis_small} %
present a qualitative visual comparison of the different explanation models on a random sample of single-class images for the ViT-B and the RN models, respectively. 
As can be seen, in terms of performance consistency and the ability to produce focused maps, LTX appears favorable compared to its alternatives.

In Fig.~\ref{fig:2_classes_vis} we further compare the ability of different ViT explanation models to produce class-specific explanations for images with two objects. Only LTX, GC, T-Attr, and GAE support class-specific maps with other baselines being class-agnostic. In LTX, class-specific explanation maps are achieved at the instance-based finetuning phase by optimizing each time for a different class. Again we see that LTX's maps are better at capturing the correct class to be explained.

Figure~\ref{fig:perturbation_tests_steps_comparison} compares LTX and T-Attr~\cite{chefer2021transformer} for the first 7 steps of the POS test. At each step, we remove an additional 10\% of the pixels reaching a 70\% pixel blackout on the right-most column.
The red square indicates the point at which the top prediction changed from the original value. We see that in both examples (the ostrich and the rabbit), the POS value is lower for LTX as it produces a mask that is more focused on the object.

Figure~\ref{fig:training_procedure_examples} exemplifies LTX's pretraining process. It shows how the explainer's mask gradually improves with  learning. Similarly, Fig.~\ref{fig:LTE_process_from_pre_LTE} demonstrates the gradual improvement obtained during LTX's finetuning step. In the left-most image, we see the explanation mask following the pretraining phase which is the starting point for the finetuning process. For this mask, the positive perturbation test halts at 20\% blackout pixels. After 4 finetuning updates (step 4), this result improved to 10\% blackout pixels.

\section{Conclusion}
\label{sec:conclusions}
We introduced Learning to Explain (LTX) - a model agnostic framework for producing explanations in which an explainer model learns to explain the predictions of a black box model via counterfactual optimization. LTX training takes place in two steps: a general pretraining phase where the explainer learns to highlight the most influential areas in the input image, followed by a per-instance finetuning stage, where the explainer model is further refined to improve its explanation of a particular sample. Our evaluations demonstrate that LTX achieves new state-of-the-art in explaining ViT and CNN models. In the future, we plan to extend the application of LTX to other domains, including natural language processing and recommender systems.

  \vspace*{0mm}

\section*{Acknowledgment}
\label{sec:ack}
This research was supported by the Israel Science  Foundation (grant No. 2243/20).

\bibliographystyle{abbrv}
\bibliography{99_egbib}

\end{document}